\documentclass[10pt, a4paper]{article}

\usepackage[]{lrec-coling2024} 
\usepackage{hyperref}       
\usepackage{url}            
\usepackage{booktabs}       
\usepackage{amsfonts}       
\usepackage{nicefrac}       
\usepackage{microtype}      
\usepackage{xcolor}         
\usepackage{graphicx}
\usepackage[nointegrals]{wasysym}
\usepackage{scalerel, amsmath}
\usepackage{adjustbox}
\usepackage{makecell} 
\usepackage{multirow}

\title{MoZIP: A Multilingual Benchmark to Evaluate Large Language Models in Intellectual Property}

\name{
	\begin{tabular}{c}
	Shiwen Ni$^{1*}$\thanks{$^{*}$Equal contribution}, 
	Minghuan Tan$^{1*}$,
	Yuelin Bai$^1$,
	Fuqiang Niu$^{2,3}$\\
	Min Yang$^{1\dagger}$\thanks{$^{\dagger}$Corresponding author}, Bowen Zhang$^{2\dagger}$, Ruifeng Xu$^4$, Xiaojun Chen$^3$,Ye Li$^1$, Jianping Fan$^1$
\end{tabular}
} 

\address{$^1$Shenzhen Institute of Advanced Technology, Chinese Academy of Sciences \\
	$^2$Shenzhen Technology University \quad
	$^3$Shenzhen University \\
	$^4$Harbin Institute of Technology, Shenzhen \\
\{sw.ni, mh.tan, yl.bai, min.yang, ye.li, jp.fan\}@siat.ac.cn,\\
nfq729@gmail.com \quad zhang\_bo\_wen@foxmail.com\\ 
xuruifeng@hit.edu.cn \quad xjchen@szu.edu.cn}

\abstract{
Large language models (LLMs) have demonstrated impressive performance in various natural language processing (NLP) tasks. However, there is limited understanding of how well LLMs perform in specific domains (e.g, the intellectual property (IP) domain). In this paper, we contribute a new benchmark, the first \textbf{M}ultilingual-\textbf{o}riented qui\textbf{Z} on \textbf{I}ntellectual \textbf{P}roperty (\textbf{MoZIP}), for the evaluation of LLMs in the IP domain. The MoZIP benchmark includes three challenging tasks: IP multiple-choice quiz (IPQuiz), IP question answering (IPQA), and patent matching (PatentMatch). In addition, we also develop a new IP-oriented multilingual large language model (called \textbf{MoZi}), which is a BLOOMZ-based model that has been supervised fine-tuned with multilingual IP-related text data. We evaluate our proposed MoZi model and four well-known LLMs (i.e., BLOOMZ, BELLE, ChatGLM and ChatGPT) on the MoZIP benchmark. Experimental results demonstrate that MoZi outperforms BLOOMZ, BELLE and ChatGLM by a noticeable margin, while it had lower scores compared with ChatGPT. Notably, the performance of current LLMs on the MoZIP benchmark has much room for improvement, and even the most powerful ChatGPT does not reach the passing level. Our source code, data, and models are  available at \url{https://github.com/AI-for-Science/MoZi}.
 \\ \newline \Keywords{ Intellectual property, benchmark, large language model, multilingual} }

\begin{document}

\maketitleabstract

\section{Introduction}
With the development of Large Language Models~(LLMs)~\cite{workshop2023bloom,openai2023gpt4,zeng2023glmb,touvron2023llama}, AI assisted agents are showing increasing abilities in understanding natural language and manipulating well-formatted text drafted by humans.
Recent studies show that LLMs supervised fine-tuned on domain-specific data achieve significant progress in a wide range of fields, such as Finance \citep{wu2023bloomberggpt}, Law \citep{LAWGPT-zh,huang2023lawyer}, Medicine \citep{huatuogpt-2023,wang2023huatuo} and Programming \citep{doi:10.1126/science.abq1158}.
As large models have exhibited remarkable versatility in our daily lives, it is crucial to effectively evaluate their abilities in handling specific tasks and identify potential shortcomings. 
Recently, several benchmarks were proposed to evaluate the large models from different perspectives such as factual consistency  \citep{laban2023llms}, question answering in the medical field \citep{singhal2022large}, the fairness of recommendation \citep{zhang2023chatgpt}, the programming ability \citep{chen2021codex}, and comprehensive evaluation in general fields \citep{huang2023ceval}.


However, we notice that in the area of AI for science, the protection and inspiration for creativity are still overlooked by the community.
Intellectual Property~(IP) has been a widely-used term to address rights in encouraging innovation and creativity.
Since 2000, the World Intellectual Property Organization~(WIPO) has established World Intellectual Property Day to ``raise awareness of how \textit{patents}, \textit{copyright}, \textit{trademarks} and \textit{designs} impact on daily life'' and ``to celebrate creativity, and the contribution made by creators and innovators to the development of economies and societies across the globe'' \citep{wipo}.
Despite the impressive capabilities of LLMs in natural language understanding and generation,
it remains a challenge to explore to what extent LLMs can understand innovative ideas, cutting-edge creations, and their protections against infringements.

As far as we are concerned, the major challenges to developing LLMs in the IP domain are two-fold. 
First, due to the wide coverage of IP rights, there is still a lack of benchmarks evaluating how LLMs understand IP-related concepts and regulations.
The major IP rights include \textit{Patents}, \textit{Trademarks}, \textit{Industrial Designs}, \textit{Geographical Indications}, \textit{Copyright} and \textit{Trade Secrets}.
Unfortunately, existing benchmarks in QA or Laws do not focus on these directions.
Second, despite the great demand for people from different professions, there are still obstacles to acquiring highly relevant information and protecting potential IPs using appropriate strategies.
According to WIPO PCT Yearly Review 2022, the top 50 PCT geographical clusters accounted for nearly
60\% of total PCT filings.
As a result, it is essential to develop a IP-oriented LLM and construct a benchmark for evaluating the performance of different LLMs. 

To make fair comparisons of LLMs over IP knowledge, we propose a multilingual benchmark called \textbf{MoZIP}~(\textbf{M}ultilingual-\textbf{o}riented qui\textbf{Z} for \textbf{I}ntellectual \textbf{P}roperty). The MoZIP benchmark consists of three datasets, IPQuiz, IPQA and PatentMatch.
Since IP rights are protected under IP systems, a prior requirement for the LLMs is domain-specific knowledge about terminologies and regulations of IP systems.
To consider the above knowledge, we first construct the IPQuiz dataset, which consists of 2000 multilingual multiple-choice questions on IP knowledge.
These questions are gathered from online IP knowledge tests from different countries and languages.
We also pay special attention to frequently asked questions on websites of IP-related organizations and agencies. These questions contain important information that IP consumers care about.
From these questions, we select 100 items to form the IPQA test set.
Among all the highlighted rights of IP, \textit{patents} encourage the development of innovations and new technologies in every field.
However, there have been numerous new patents being submitted each day that searching or indexing similar patents has been a challenge for current systems.
We collect patent documents covering different languages across the globe to help the model to learn how patents are drafted and innovations are described.
Based on this, we further construct the PatentMatch dataset to test how models may differentiate patents based on the descriptions. 

In this paper, we also develop our IP-oriented LLM \textbf{MoZi}.
The model is finetuned from BLOOMZ-MT-7B through three stages.
For the first stage, we use 24 million official patent documents to make the model aware of how patents are usually crafted.
For the second stage, we conduct 3 million instruction finetuning using various instruction types across multiple fields.
In the last stage, 58k IP instruction fine-tuning data constructed by ourselves is used to enable the model to learn about IP knowledge.
We conducted experiments based on five LLMs of MoZi, ChatGPT, ChatGLM, BELLE, and BLOOMZ and evaluated each model on our benchmark \textbf{MoZIP}. Our experiments show that ChatGPT performs best overall, followed by the other 6-7b parameter number models in which MoZi performs best. Overall, there is still much room for improvement in the current LLMs' underperformance on MoZIP.
The contributions of this paper to the community are summarized as follows:

\begin{itemize}
	\item This paper presents \textbf{MoZIP} the first IP benchmark covering nine languages for evaluating the capabilities of large language models in the IP domain. 
	\item In this work, we propose the first IP-oriented multilingual large language model \textbf{MoZi}, and experimental results on the MoZIP benchmark show that \textbf{MoZi} performs the best among models of the same parameter level.
	\item We conducted a comprehensive evaluation using five LLMs on the MoZIP benchmark, and the experimental results show the challenge of the MoZIP benchmark and illustrate the deficiencies of LLMs in the IP field at this stage.
	\item In order to contribute to the development of LLMs in intellectual property, we have made available source code, \textbf{MoZIP} benchmark, instruction fine-tuning data, and \textbf{MoZi} model.
\end{itemize}

\section{Primary data collection}
\begin{figure*}[ht]
	\centering
	\includegraphics[width=1\linewidth]{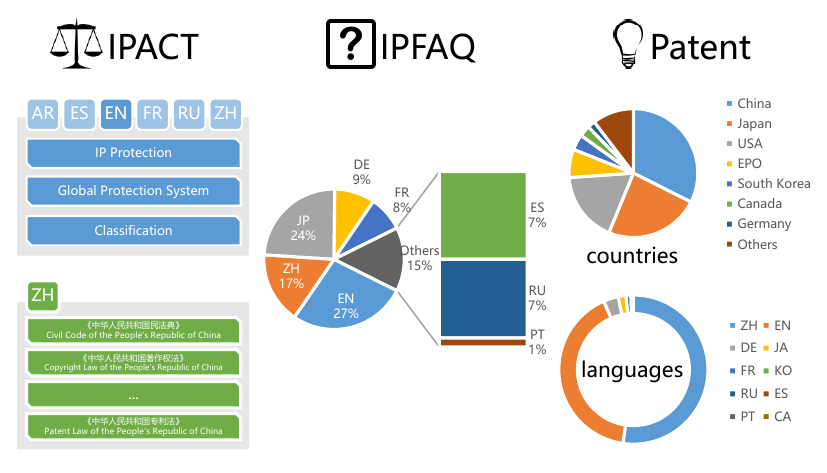}
	\caption{Statistics and distribution of data. ZH-Chinese, EN-English, DE-German, JA-Japanese, FR-French, KO-Korean, RU-Russian, ES-Spanish, PT-Portuguese, CA-Catalan.} 
	\label{fig:data}
\end{figure*}

To facilitate the construction of an IP-oriented benchmark, we collect different categories of data from multiple resources, including IP acts, patents as well as frequently asked questions related to IP.
These data can be used as instructions for fine-tuning to inject IP-related knowledge into LLMs.

\textbf{IPACT} To protect IP rights, there has been plenty of legal information on intellectual property available online, including IP laws and regulations, WIPO-administered and IP related treaties, and leading judicial decisions on IP. This part of the data is basically objective knowledge.
LLMs need to access such resources to acquire knowledge about mechanisms of IP systems.

\textbf{IPFAQ} Frequently Asked Questions~(FAQs) related to IP usually contain basic knowledge about IP rights and important information that IP customers care most about but likely to misunderstand.
We crawl FAQs from worldwide websites of IP organizations and agencies. Their answers are officially provided with guaranteed reliability and accuracy. The dataset therefore contains questions asked by people from different countries and answers given online.

\textbf{Patent} “A patent is an exclusive right granted for an invention, which is a product or a process that provides.” \cite{wipo}
Generally speaking, a patent needs to propose a new way of doing something, or offer a new technical solution to a problem. 
In order to obtain a patent, one must reveal technical details about their invention in a patent application that is made available to the public.

We gather patent documents from a database of CNIPA\footnote{The Patent data is crawled from \url{http://www.szxyd.org.cn} before March 15, 2023.}.
Figure~\ref{fig:data} shows an example of a patent document.
Each patent document contains different fields of data, like \textit{country of the patent}, \textit{patent type}, \textit{public number}, \textit{claims}, \textit{International Patent Classification (IPC) information} and \textit{description}.
The patents may be written in different languages regardless of the \textit{country of the patent}.
We use the \textit{claims} field and the \textit{description} field as language indicators.
If the two fields are empty~(about half of the Patent data has this issue), the patent is not used for language detection.
We adopt a language detector from FastText \cite{joulin2016fasttext,joulin2016bag} to categorize the patents into different language groups.
The right of Figure~\ref{fig:data} shows the statistics for country and language distributions for all used patents.

\section{Benchmark} 
\begin{table}[ht]
	\centering
	\setlength\tabcolsep{3pt}
	\begin{tabular}{lccc}
		\toprule
		Dataset &\multicolumn{1}{c}{IPQuiz} & \multicolumn{1}{c}{IPQA}& \multicolumn{1}{c}{PatentMatch} \\\midrule
		Types & \multicolumn{1}{c}{M-C} &\multicolumn{1}{c}{Generation}  &\multicolumn{1}{c}{M-C} \\\midrule
		EN & 834 & 35 &500 \\
		ZH & 564 & 35 &500 \\
		XL & 623 & 30 &- \\\midrule
		Size & \multicolumn{1}{c}{2021} & \multicolumn{1}{c}{100}& \multicolumn{1}{c}{1000} \\\midrule
		\# languages & \multicolumn{1}{c}{7} & \multicolumn{1}{c}{7}& \multicolumn{1}{c}{2} \\
		\bottomrule \\
	\end{tabular}
	\caption{Statistics for each dataset in MoZIP benchmark. IPQuiz-XL includes DE, ES, JP, KO and PT. IPQA-XL includes ES, JP, DE, FR, RU. M-C: Multiple-choice.}
	\label{tab:mozip}
\end{table}

The \textbf{MoZIP} benchmark contains \textbf{IPQuiz}, \textbf{IPQA} and \textbf{PatentMatch}, see Table~\ref{tab:mozip}.

\begin{figure*}[t]
	\centering
	\includegraphics[width=1\linewidth]{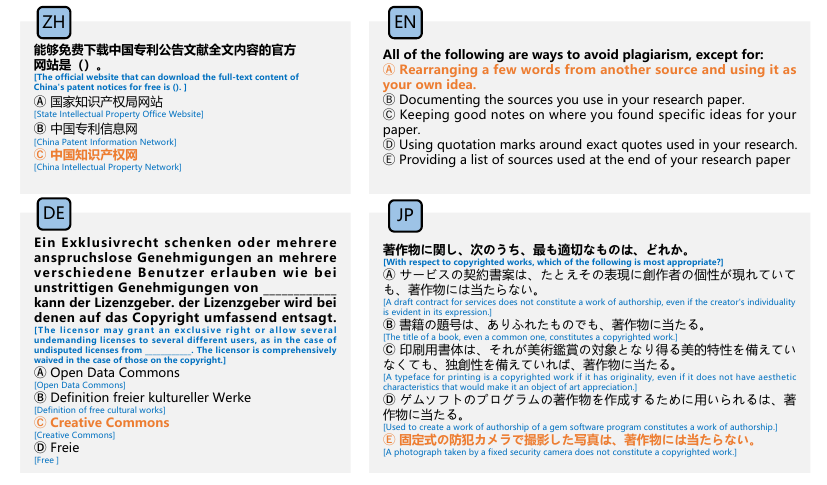}
	\caption{Examples of questions in IPQuiz. The words in blue below non-English content are the corresponding English translations.} 
	\label{fig:ipquiz}
\end{figure*}

\textbf{IPQuiz}\footnote{\url{https://huggingface.co/datasets/BNNT/IPQuiz}}  We construct the IPQuiz dataset to evaluate if models understand IP-related concepts and regulations.
IPQuiz is a multiple-choice question answering dataset gathered from publicly accessible websites of various languages across the globe.
For each question, the model needs to choose a response from a list of candidates, see Figure~\ref{fig:ipquiz}.
The dataset is keeping increasing in size and coverage of languages.
We will host the dataset on HuggingFace's \textit{datasets} website.
For the initial version of IPQuiz, it contains 2k questions with seven languages.
Figure~\ref{fig:data} shows the distribution of languages for the initial version of the dataset.

\begin{figure}[t]
	\centering
	\includegraphics[width=1\linewidth]{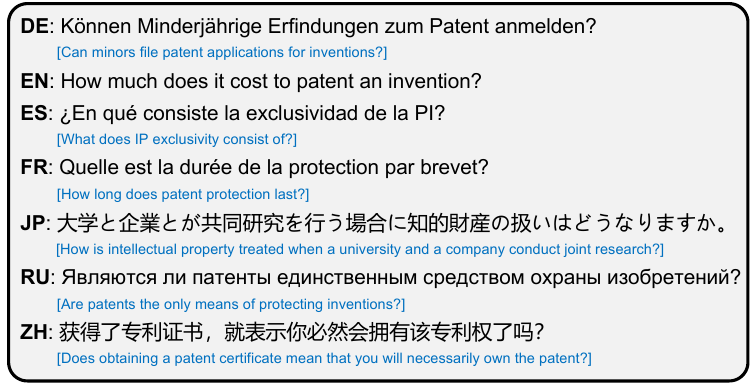}
	\caption{Examples of seven language questions in IPQA dataset. The words in blue below non-English content are the corresponding English translations.} 
	\label{fig:ipq}
\end{figure}

\textbf{IPQA}\footnote{\url{https://huggingface.co/datasets/BNNT/IPQA}} We reserve 100 questions from \textbf{IPFAQ} as test set to evaluate how LLMs understand IP-related questions. The IPQA contains questions in seven languages, and the 100 data items include 35 each in Chinese and English, and 6 each in Spanish, Japanese, German, French, and Russian. The specific IP-questions are shown in Figure \ref{fig:ipq}.
Their responses are further judged by human annotators to compare the quality.

\begin{figure*}[t]
	\centering
	\includegraphics[width=\linewidth]{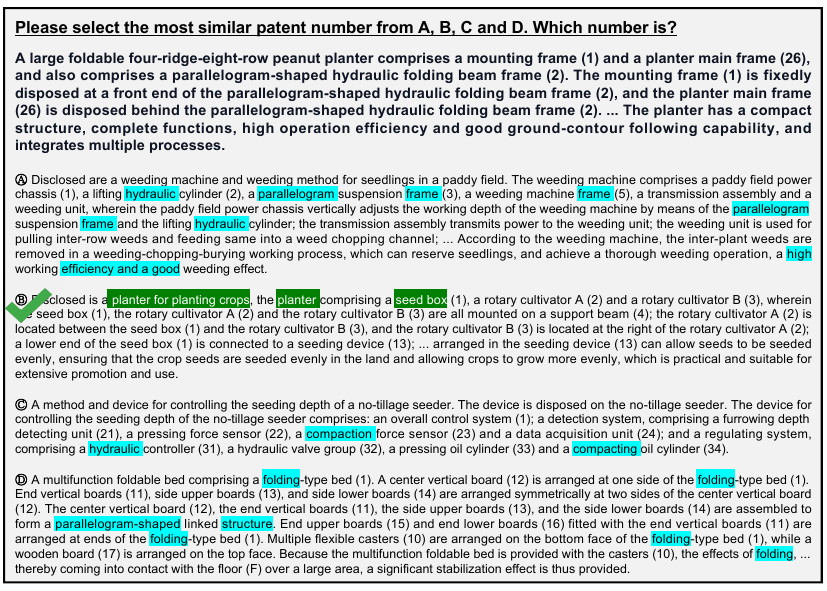}
	\caption{An example in PatentMatch. The texts with blue color are overlappings between the source patent and each candidate patent. However, the texts with green background color are the key information why the two patents match each other.} 
	\label{fig:patentmatch}
\end{figure*}

\textbf{PatentMatch} \footnote{\url{https://huggingface.co/datasets/BNNT/PatentMatch}}
We construct PatentMatch to assess whether the model truly comprehends the inventions described in patent documents and accurately differentiates between different patents. We first create a parallel dataset of 250k patents granted by WIPO from 2010 to 2022 collected from Google Public Datasets \footnote{\url{https://console.cloud.google.com/marketplace/product/google_patents_public_datasets/google-patents-public-data}}, with both Chinese and English abstracts extracted from the original documents. These contents are written by the patent applicants, ensuring the accuracy of the descriptions. 
We construct this sub-task following the steps below: 
\begin{itemize}
	\item First, we leverage the Pyserini toolkit \cite{Lin_etal_SIGIR2021_Pyserini} to build a BM25 database. And we use OpenAI API (text-embedding-ada-002 model) \footnote{ \url{https://platform.openai.com/docs/guides/embeddings}} to build a dense vector database based on Pinecone\footnote{\url{https://www.pinecone.io/}} with cosine similarity metric. 
	\item Then, we select 500 patents across various technical domains based on 8 sections from the IPC classification system to construct 1000 multiple-choice questions(500 questions in each language). Next, we use the abstracts of these patents as queries to retrieve the top-k results from both the BM25 database and the dense vector database.
	\item As our setting is to select the most relevant patent from the options, we follow the following steps to construct these choices. (1) To determine the accurate answer, we choose patents that belong to the same IPC subgroup as the patent mentioned in the question. We then make sure these patents have a low BM25 ranking but a high ranking in vector-based assessments. This setup allows us to assess whether the model can effectively recognize similar patents even when there are fewer word overlaps. 
	(2) We choose the remaining options based on patents with high rankings in BM25 but low rankings in the vector-based assessment.  Additionally, we ensure that these patents have a different IPC classification code with patent in the question. 
\end{itemize}


Human evaluators then perform manual verification to validate the quality of the constructed questions. We made necessary modifications or deletions based on their feedback, resulting in a final set of 500 PatentMatch questions in both English and Chinese.
\section{The Proposed MoZi Model}
\begin{figure*}[t]
	\centering
	\includegraphics[width=1\linewidth]{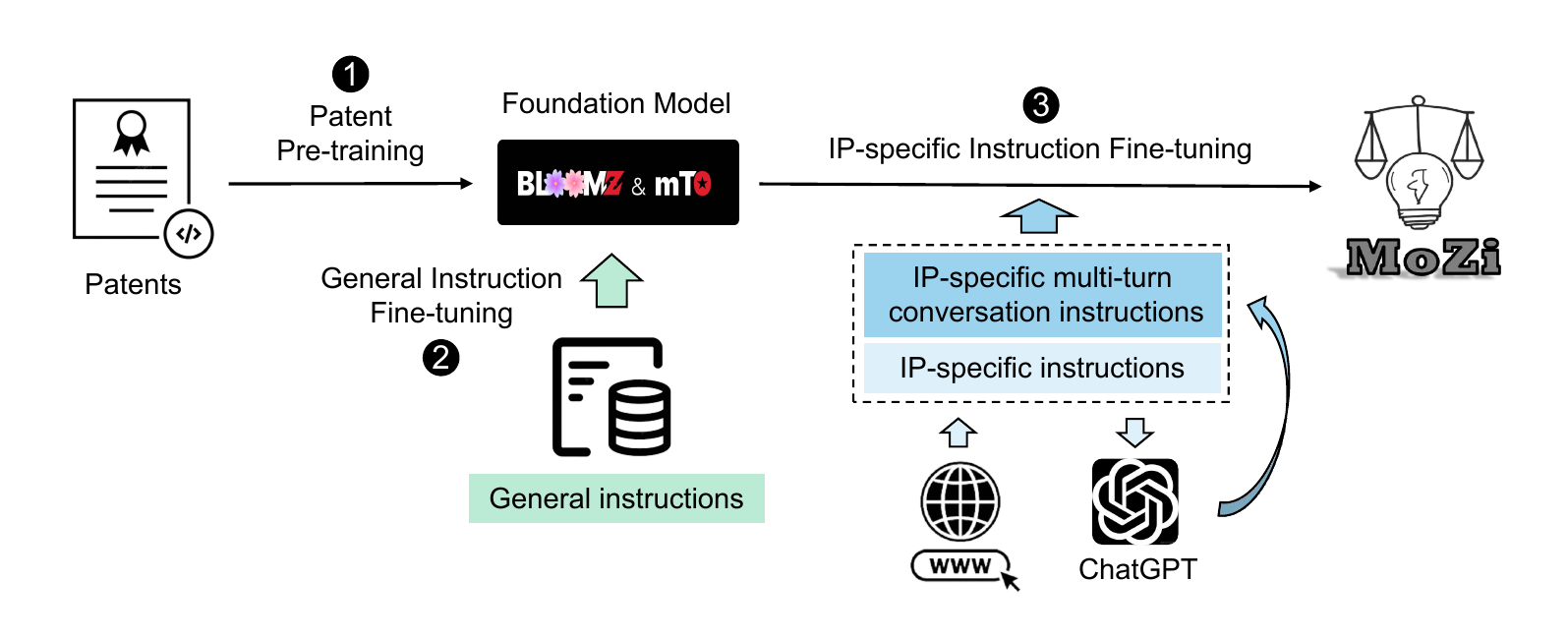}
	\caption{Schematic of our proposed IP-oriented multilingual large language model \textbf{MoZi}.} 
	\label{fig:model}
\end{figure*}

As depicted in Figure \ref{fig:model}, we conduct three stages of fine-tuning on an IP-oriented language model known as \textbf{MoZi}, which is built upon the BLOOMZ-MT-7B model. These stages entail the utilization of the data gathered in Section 2. Due to the substantial amount of Patent data available, we choose to employ it in the initial stage. Following that, we proceed with two additional stages of instruction finetuning, incorporating both general-domain instructions and IP-domain-specific instructions.

\textbf{Patent Pre-training} Patent documents contain detailed descriptions and instructions about the inventions.
However, in practice, most models struggle in differentiating patents or retrieving similar patents given the large amount of candidates and high overlapping of text.
To strengthen abilities in this direction, we use the Patent Data to train our model \textbf{MoZi}.
Our model is initialized from a publicly released checkpoint\footnote{\url{https://huggingface.co/bigscience/bloomz-7b1-mt}} of BLOOMZ-MT, which is a multilingual model finetuned on xP3mt~\cite{muennighoff2022crosslingual}.
We construct a training case from each patent document.
The context includes the patent's title, abstract, claims and description.

\textbf{General Instruction Finetuning}\footnote{\url{https://huggingface.co/datasets/BNNT/mozi_general_instructions_3m}}
In total, we used 3,025,600 general instructions for the first stage of our proprietary pre-trained model. The 3m instruction data come from several public datasets, including 1) BELLE Chinese general instructions, 2) Alpaca-gpt4 English general instructions, 3) BELLE Chinese general conversation instructions, 4) Sharegpt-vicuna English general conversation instructions, and 5) GuanacoDataset Chinese, English, and Japanese general instructions.

\textbf{IP-specific Instruction Finetuning}\footnote{\url{https://huggingface.co/datasets/BNNT/mozi_IP_instructions}} 
For the model after fine-tuning the generic instruction, we perform the second stage of IP-specific Instruction Finetuning upon it. The fine-tuned data in this phase consisted of a total of 58,874 items, including multilingual Q\&A datasets we collected from the IPFAQ dataset, Chinese IP-related law articles in IPACT, and Chinese multi-turn conversation data in the IP domain generated by ChatGPT. Let's think of each law as an instruction, where the input is the name of the law and the output is the specifics of the law. ChatGPT generates a prompt for a multiple round conversation: "\textit{The following is a conversation between a user and a patented AI assistant. The user and the patented AI assistant are having a conversation around this topic: [ seed question ]. The user utterance begins with human and the AI assistant utterance begins with assistant. The user asks relevant questions about the topic in question or about previous conversations. When they have no more questions, the user will stop the conversation. The AI assistant tries not to ask questions.}"
\begin{table*}[t]
	\centering
	\setlength\tabcolsep{8pt} 
	\begin{tabular}{lccccc}
		\toprule
		\multirow{2}{*}{Dataset} & \multicolumn{5}{c}{Model}                                   \\
		\cmidrule{2-6} 
		& MoZi-7b (ours) & BLOOMZ-7b & BELLE-7b & ChatGLM-6b & ChatGPT \\ 
		\midrule
		IPQuiz-EN                  & 41.5          & 29.3      & \underline{42.7}     & 38.9       & \textbf{60.8}    \\
		IPQuiz-ZH                  & \underline{39.2}          & 29.1      & 38.6     & 31.7       & \textbf{45.8}    \\
		IPQuiz-XL                  & \underline{37.4}          & 29.4      & 36.1     & 30.5       & \textbf{42.2}    \\\midrule
		Average                  & \underline{39.4}          & 29.3      & 39.1     & 33.7       & \textbf{49.6} \\ \bottomrule\\
	\end{tabular}
	\caption{Performance of all used models on IPQuiz.}
	\label{tab:ipquiz}
\end{table*}

\begin{figure*}[t]
	\centering
	\includegraphics[width=1\linewidth]{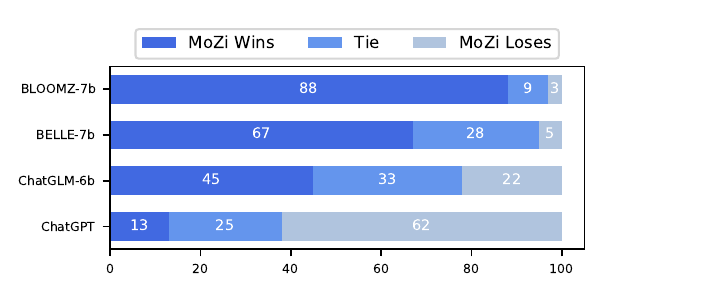}
	\caption{The number of questions on IPQA that MoZi wins, ties or loses.} 
	\label{fig:comparison}
\end{figure*}
\section{Experiments}

\subsection{Training Details}
Our MoZi-7b model was implemented in PyTorch using the \textit{transformer} and \textit{deepspeed} packages and with Bloomz-7b1-mt \cite{workshop2023bloom} as the Foundation model. We used ZeRO-3 \cite{rajbhandari2020zero} to distribute the model over 8 × A100 GPUs for training. During supervised fine-tuning, we set the learning rate, batch size, and maximum token length to 5e-6, 64, and 2048, respectively. Moreover, weight\_decay is set to 0.0001 and num\_warmup\_steps is set to 100. The training epochs for the pre-training phase are 1 time, the general instruction fine-tuning phase is 2 times, and the IP instruction fine-tuning stage is 4 times.
\subsection{Baselines}
In addition to evaluating the proposed \textbf{MoZi-7b} on the MoZIP benchmark, we evaluated the following baseline models:
\begin{itemize}
	\item \textbf{ChatGPT \footnote{\url{https://openai.com/blog/chatgpt}}} A powerful large language model developed by OpenAI, which is based on the \textit{gpt-3.5-turbo} model. We used the API provided by openai for the evaluation.
	\item \textbf{ChatGLM-6b \cite{du2022glm,zeng2022glm}} An open source, Chinese-English bilingual large language model proposed by Tsinghua University, based on the general language model (GLM) architecture.
	\item \textbf{BELLE-7b \cite{BELLE,belle2023exploring}} It is a Bloomz-7b1-mt fine-tuned model that combines 2 million Chinese data and 50,000 English data from the open source Stanford-Alpaca, which makes it have excellent Chinese comprehension and response generation capabilities.
	\item \textbf{BLOOMZ-7b \cite{workshop2023bloom,muennighoff2022crosslingual}} A LLM obtained by fine-tuning the BLOOM and mT5 pre-trained multilingual models on a cross-language task mixture (xP3) with the ability to generalize across languages to unseen tasks and languages.
\end{itemize}

The LLaMA family of large language models performs adequately only in English, and its multilingual capability is limited. Consequently, we opted not to employ the LLaMA family of models for measurement purposes.

\subsection{Experimental Results}

\textbf{IPQuiz} consists of 2k multiple-choice questions about IP knowledge covering 7 languages.
We construct this prompt (“\textit{Please give the correct option for the following question}”) to LLMs and collect generated responses from them. 
We used regularity and manual verification to determine which candidate was selected as the answer.
Then we compute answer accuracy over the dataset. The experimental results are shown in Table \ref{tab:ipquiz}, ChatGPT performs the best on IPQuiz-zh, IPQuiz-en and IPQuiz-xl because of its parameter level and the amount of training data. With the exception of ChatGPT, our MoZi-7b had the best average accuracy of 39.4\% among models with approximately 7b number of parameters. Compared to the foundation model we used, BLOOMZ-7b, our MoZi-7b improved on average by 10.1\% on IPQuiz tasks in various languages, which illustrates the significant effect of further pre-training and instruction fine-tuning. Overall, none of the scores exceeded 60\%, except for ChatGPT, which reached 60.8\% on IPQuiz-en. This shows that current LLMs almost always fail in performance on our proposed IPQuiz dataset. There is a lot of room for improving the performance of LLMs in the IP domain at this stage.

\begin{figure*}[t]
	\centering
	\includegraphics[width=1.0\linewidth]{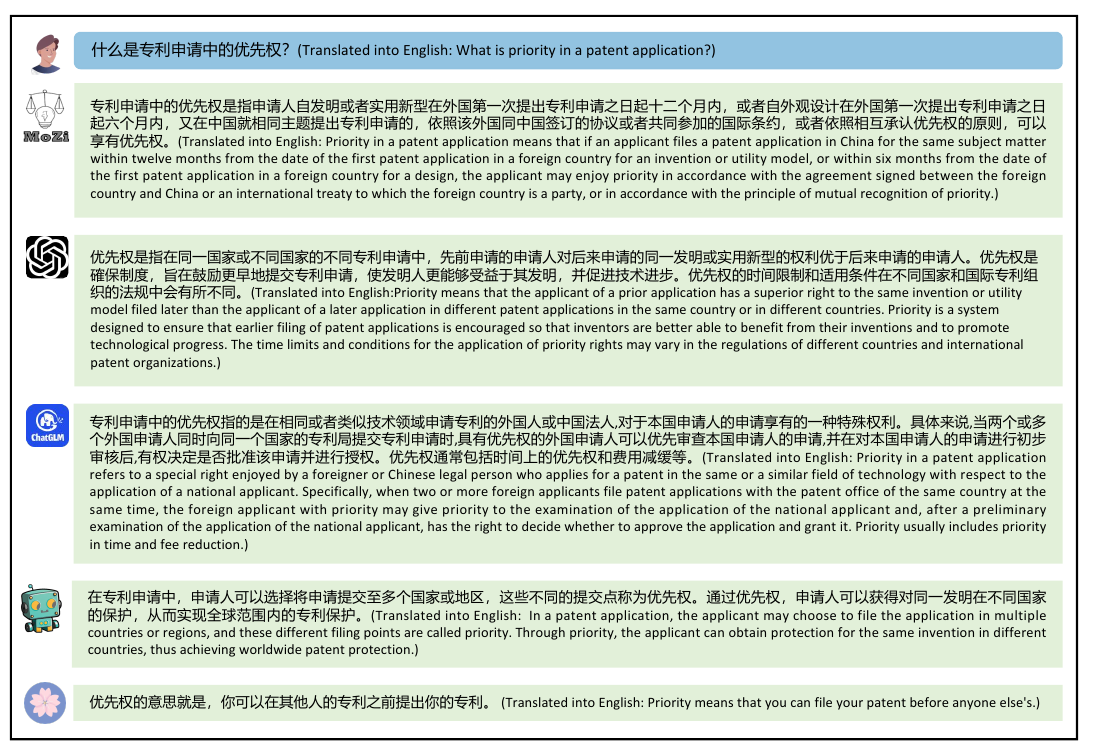}
	\caption{Comparisions for answers generated by diffenrent models for a question in IPQA, the models from top to bottom are MoZi-7b, ChatGPT, ChatGLM-6b, BELLE-7b and BLOOMZ-7b.} 
	\label{fig:ipqa}
\end{figure*}
\begin{table*}[ht]
	\centering
	\setlength\tabcolsep{8pt} 
	\begin{tabular}{lccccc}
		\toprule
		\multirow{2}{*}{Dataset} & \multicolumn{5}{c}{Model}                                   \\ \cmidrule{2-6} 
		& MoZi-7b (ours) & BLOOMZ-7b & BELLE-7b & ChatGLM-6b & ChatGPT \\ \midrule
		EN	&\underline{25.8} &	24.0	&14.2&	23.8&	\textbf{34.6}   \\
		ZH	&\underline{29.0}& 	24.2&	26.4&	27.2&	\textbf{43.0}    \\
		\midrule
		Average           &       \underline{27.4} &	24.1&	20.3	&25.5	&\textbf{38.8} \\ \bottomrule\\
	\end{tabular}
	\caption{Evaluation results for all models over PatentMatch.}
	\label{tab:patent-matching}
\end{table*}

\textbf{IPQA} We provide the annotators with a question and two potential answers generated by different models. The annotators were then tasked with indicating which answer they thought was superior, or whether there was a significant difference between the two answers. If there is no distinction then label the tie.
We assess inter-annotator agreement using tie-discounted accuracy. In this method, we award one point when both annotators agree on a label, half a point when either annotator (but not both) labels a tie, and no points when there is no agreement. We compared two different human annotators and obtained a consistent score of 81\%. Despite the presence of some subjective elements in this task, the agreement among human annotators is reasonably good.

The results of the experiments on IPQA are shown in Figure \ref{fig:comparison}, and are ChatGPT, MoZi-7b, ChatGLM-6b, BELLE-7b, and BLOOMZ-7b in descending order of performance results. Our MoZi-7b model defeated the original foundation model, BLOOMZ-7b, a total of 88 times, losing only 3 times. Also both BLOOM-based models, our MoZi-7b beat BELLE-7b a total of 67 times, losing only 5 times and tying 28 times. ChatGLM-6b has performed well in IPQA, losing to MoZi-7b only 45 times, winning over MoZi-7b 22 times and tying 33 times. Finally comparing to the powerful ChatGPT, our MoZi-7b also beat it 13 times and tied it 25 times.

In addition, Figure \ref{fig:ipqa} is a specific IPQA example that we show. From top to bottom are the human questions, the answer of MoZi-7b, the answer of ChatGPT, the answer of ChatGLM-6b, the answer of BELLE-7b and the answer of BLOOMZ-7b, respectively. MoZi-7b has the most accurate answer.

\textbf{PatentMatch} The experimental results for the five LLMs on the PatentMatch task are shown in Table \ref{tab:patent-matching}.
ChatGPT still performs the best. 
However, we found that ChatGPT only got 43.0\% and 34.6\% accuracy on PatentMatch-zh and PatentMatch-en, respectively. The mean scores of the four models MoZi-7b, ChatGLM-6b, BELLE-7b and BLOOMZ-7b were 27.4\%, 24.1\%, 20.3\% and 25.5\%, respectively. Our MoZi is also better than other LLMs in its class in both PatentMatch-zh and PatentMatch-en. 
That is, the average score of the other four models except ChatGPT is useless to exceed 30\%. We also found that BELLE-7b's score in PatentMatch-en was even only 14.2\% far worse than a blind guess. We believe that the reason why all models perform so poorly may be that patent data are too rare and that LLMs have more difficulty understanding long texts. The input of the models in the PatentMatch task is basically more than 1,000 tokens. From the experimental results, it still faces a great challenge for LLMs to process long texts in specialized fields.

\section{Conclusion}
We introduce \textbf{MoZIP}, the inaugural multilingual benchmark for evaluating Language Models (LLMs) in the field of Intellectual Property (IP). MoZIP encompasses a wide range of real-world objective knowledge and various types of user questions. It comprises three distinct datasets: \textit{IPQuiz}, \textit{IPQA}, and \textit{PatentMatch}, which collectively cover nine languages. As part of this research, we present \textbf{MoZi}, the first IP-oriented Multilingual Large Language Model. We evaluate five models, namely MoZi-7b, ChatGPT, ChatGLM-6b, BELLE-7b, and BLOOMZ-7b, using the MoZIP benchmark. The experimental findings highlight the challenging nature of the MoZIP benchmark and the current deficiency of IP-related knowledge among LLMs.
To facilitate research within the recommendation community, we release the source code, benchmark datasets, instruction fine-tuning data, and the MoZi model. Our aim is for MoZIP to serve as a standardized benchmark for evaluating LLMs in the IP domain. Looking ahead, we plan to develop a more comprehensive dataset that includes a greater number of minor languages. 
\vspace{-0.5em}
\section{Ethics Statement}
The sources of our data are publicly available URLs on the Internet, URLs that can be accessed without registering for an account. And the data does not involve personal privacy. The data are collected legally. The sources of all our data are open and transparent. The purpose of the data use is to evaluate and improve the capabilities of large-scale language models in the field of intellectual property.

\section{Acknowledgements}
This research was funded by the State Sponsored Postdoctoral Researcher Program of China (GZC20232873) and China Postdoctoral Science Foundation (2023M733654).
\nocite{*}
\section{Bibliographical References}\label{sec:reference}
\bibliographystyle{lrec-coling2024-natbib}
\bibliography{lrec-coling2024-example}

\end{document}